\documentclass[10pt, journal]{IEEEtran}
\ifCLASSINFOpdf
\else
\fi

\usepackage{amsmath}
\usepackage{epsfig}
\usepackage{multirow}
\usepackage{tabularx}
\usepackage{color}
\usepackage{algorithmicx}
\usepackage{algpseudocode}
\usepackage{bm}
\usepackage{amsfonts}
\usepackage{graphics}
\usepackage{url}
\usepackage{soul}

\soulregister{\cite}7
\soulregister{\citep}7
\soulregister{\citet}7
\soulregister{\ref}7
\soulregister{\pageref}7
\soulregister{\uppercase\expandafter}7
\sethlcolor{white}

\begin{document}

\title{An Active Sense and Avoid System for Flying Robots in Dynamic Environments}

\author{Gang~Chen, Wei~Dong, Xinjun~Sheng, ~\IEEEmembership{Member,~IEEE,}, Xiangyang Zhu, ~\IEEEmembership{Member,~IEEE,} and Han Ding, ~\IEEEmembership{Member,~IEEE,}
\thanks{Gang~Chen, Wei~Dong, Xinjun~Sheng, Xiangyang Zhu and Han Ding are with
        the State Key Laboratory of Mechanical System and Vibration,
        School of Mechanical Engineering, Shanghai Jiaotong University,
        Shanghai, 200240, China.}
\thanks{Corresponding author: Wei~Dong, E-mail: dr.dongwei@sjtu.edu.cn.
}
        }

%

\maketitle

\begin{abstract}
\hl{This paper investigates a novel active-sensing-based obstacle avoidance paradigm for flying robots in dynamic environments.}
Instead of fusing multiple sensors to enlarge the field of view (FOV), we introduce an alternative approach that utilizes a stereo camera with an independent rotational DOF to sense the obstacles actively.
In particular, the sensing direction is planned heuristically by multiple objectives, including tracking dynamic obstacles, observing the heading direction, and exploring the previously unseen area.
With the sensing result, a flight path is then planned based on real-time sampling and uncertainty-aware collision checking in the state space, which constitutes an active sense and avoid (ASAA) system.
\hl{Experiments in both simulation and the real world demonstrate that this system can well cope with dynamic obstacles and abrupt goal direction changes.
Since only one stereo camera is utilized, this system provides a low-cost and effective approach to overcome the FOV limitation in visual navigation.}
\end{abstract}

\begin{IEEEkeywords}
Collision Avoidance, Active Vision, Motion and Path Planning, Visual Navigation
\end{IEEEkeywords}

\IEEEpeerreviewmaketitle

\section{Introduction}
Obstacle avoidance is one of the fundamental abilities of flying robots.
Fast and precise sensing \hl{of} obstacles is required in the first place to achieve this ability in unknown environments. When potential dynamic obstacles, such as pedestrians and other robots, exist, the situation becomes more complex, and the sensing is more significant.

\hl{Multiple types of sensors can aid the sensing process. The stereo camera is a cheap and light-weight sensor that can provide both depth and color information in high resolution and is thus widely utilized on flying robots.} However, a nonnegligible limitation of this sensor is that the FOV is usually narrow. To ensure safety, an intuitive approach is to generate path candidates only inside of the FOV \cite{AggressiveSample}, \cite{OpticalFlowDepth}, while the outside area is considered to be invalid, which leads to a heavily constrained planning effectiveness. In contrast, building an obstacle map could enlarge the valid area for path planning \cite{Ringbuffer}, \cite{OctoMap}. However, the free space in the map that is not covered by FOV currently could still be insecure because of the existence of dynamic obstacles.

To better adapt to the dynamic environment, one solution is to mount multiple stereo cameras that can cover 360 degrees \cite{FPGAOmni}, which enables real-time detection of obstacles from an arbitrary direction. Nevertheless, an obvious disadvantage is that the weight of the sensors and the required computing resource would be multiplied. Another straightforward solution is to control the fuselage's yaw angle to make the camera heading direction follow flight velocity direction \cite{DAAdynamic}, \cite{SoftYawConstrain}, which ensures the timely update of the area along the planned path. However, for many flying robots, such as quadrotors, yaw rotation is coupled with linear motion \cite{QuadModel}, and thus this approach would sacrifice flight performance. Furthermore, dynamic obstacles could still come from the blind area and cause a severe crash.

%
%


\begin{figure}
\centerline{\psfig{figure=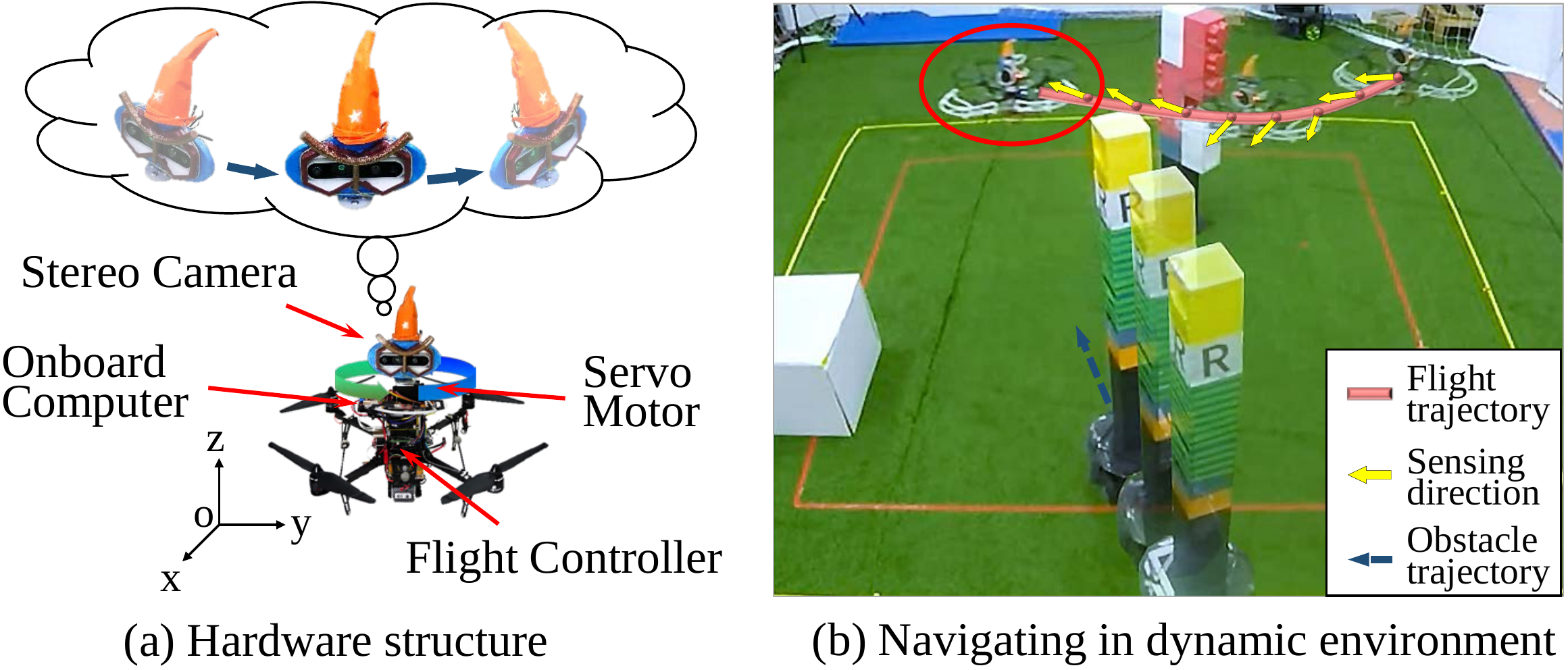,width=3.3in}}
\caption{Our flying robot in a dynamic environment. (a) presents the hardware structure of our system. (b) illustrates a scenario where the flying robot maneuvers through the gap between a dynamic obstacle and a static obstacle.}
\label{MAV}
\end{figure}

Inspired by the owl, we propose a new paradigm to tackle obstacle avoidance in dynamic environments with active stereo vision.
Although owls are unable to move their eyeballs in any direction (similar to stereo cameras), they have a very flexible neck that can swivel up to $270^\circ$, \hl{which enables them to observe rapidly even behind without relocating the torso \cite{Owl}}.
In our work, a servo motor and a stereo camera are mounted on a quadrotor to play the roles of the ``neck'' and the ``head'', as is illustrated in Fig. \ref{MAV}(a). The head is only a light stereo camera and hence can swivel much faster than the body while bringing very little influence to flight performance.
We estimate the sense update degree (SUD) of each direction and plan the head rotation angle with multiple objectives. Tracking and predicting the trajectories of dynamic obstacles are considered to adapt to dynamic environments.
Based on the sensing result, a collision-free path for the flying robot is planned in real-time through sampling in state space.
Altogether, this system is called an ASAA system. To the best of the authors’ knowledge, this is the first system that applies active stereo vision to realize obstacle avoidance for flying robots.



The main contributions of this paper are as follows:
\begin{enumerate}
\item \hl{A novel active-sensing-based obstacle avoidance paradigm that can overcome the FOV limitation in visual navigation.}
\item \hl{A sensor planning approach based on multi-objective optimization, which enhances the active sensing quality in dynamic environments.}
\item \hl{Improvements on a sampling-based path planner by considering the state estimation uncertainty from sensing.}
\end{enumerate}

\section{Related Work}
Obstacle avoidance is a vital capability for flying robots. Sensing and path planning are two essential components of obstacle avoidance.
Multiple types of sensors can be used for sensing, such as lidar, radar, monocular camera, and stereo camera. Among these sensors, lidar and stereo camera can provide high resolution position information of obstacles and thus are widely used \cite{Lidar1}, \cite{Lidar2}, \cite{FPGAOmni}, \cite{ReactiveStereoETH}. Lidar could have a detection range up to 360 degrees but is often heavy and expensive, and is unable to provide color information. In comparison, the stereo camera is lighter and can output both position and color information. Nevertheless, the limitation of the narrow FOV is nonnegligible. Obstacles outside of the FOV could be dangerous to the flight, especially in dynamic environments. Although some researchers have studied special stereo cameras with omnidirectional vision \cite{SingleCameraToFormStereoLens}, \cite{SwivelingSingleCameraToFormStereo}, these cameras are not popularized because of limitations like low precision and large size.

The fusion of multiple stereo cameras \cite{FPGAOmni}, or multiple types of sensors \cite{MultipleSensors} are usually adopted to enlarge the FOV. This approach is practical but significantly increases the weight and computation cost of the system.
An alternative approach is to adopt active vision \cite{ActiveVision} to change the FOV to the direction that benefits more to flight safety.
One way to realize this kind of FOV changing on a flying robot is to use a fixed camera mounted in the front and rotate the fuselage yaw direction. Usually, the yaw direction follows the currently planned velocity direction to align the FOV along the flight path \cite{DAAdynamic}, \cite{SoftYawConstrain}. In accordance with the dynamic property of flying robots like multirotors \cite{QuadModel}, yaw rotation is strongly coupled with the linear motion, and thus obvious control error on the linear motion can be caused by rotating the yaw angle, especially when large curvature occurs on the flight path and the yaw angle has to be rotated fast.
To reduce this control error, \cite{SoftYawConstrain} applies additional constraints in the flight path planning phase to make the flight path's curvature smaller.
In static environments, the way of aligning the FOV along the flight path is sufficient.
However, when dynamic obstacles exist, more observation towards dynamic obstacles, which could come from any direction, is required to predict their future trajectories for collision avoidance. The FOV direction is supposed to change frequently and rapidly to achieve observation towards more directions. A more effective approach is still required.

With the observation data from sensors, a collision-free path can be planned to realize obstacle avoidance. Some works directly use point cloud \cite{AggressiveSample}, or depth image \cite{DepthReactive1}, \cite{DepthReactive2} in the FOV to plan the path, which reacts fast but the planning range is constrained in the FOV. Others prefer building a map to store the previously observed result and then plan in the map through search-based method \cite{ZouDynamic}, \cite{ETHOldMap}, optimization-based \cite{Ringbuffer} method, or the combination of both \cite{ZhouBoYu}.
Compared to the methods that only use the sensing result in FOV, map-based methods could enlarge the valid area to search for safe paths. The most popular map is the voxel map, such as Octomap \cite{OctoMap} and circular map \cite{Ringbuffer}. These voxel maps are built upon the probability accumulation of occupancy status at discrete positions in the 3D space and hence are suitable for static environments. Regarding the dynamic obstacles, the response of this accumulation paradigm is not fast enough, and lots of noise would occur on the map.
The work in \cite{DAAdynamic} tries to decrease the response time and reduce noise by removing and re-adding the voxels inside of FOV. 

To better fit the dynamic environment, a sense and avoid (SAA) system \cite{SAA} that can predict the future trajectories of dynamic obstacles and plan a flight path to avoid collisions is required.
In a relatively high flight space for large unmanned aerial vehicles, the environment is usually open, and dynamic obstacles are other aerial vehicles. The SAA system mainly concerns avoiding collision towards other aerial vehicles \cite{PoseEstimateSAA}.
In a near-ground environment for the flying robot like ours, the situation is more complicated. Various kinds of dynamic obstacles could appear in addition to static obstacles. Only a few works have tackled this issue.
\cite{Dynamic2020} considers moving obstacles like pedestrians in the front depth view and use a chance-constrained model predictive controller to realize fast and collision-free flight.
The work in \cite{MapPredictionSAA} applies a Bin-Occupancy filter to track dynamic obstacles in a voxel map. The flight path is planned by optimizing the real-time control commands with a low probability of collision in the next N steps. Due to the complexity caused by dynamic obstacles, there is still huge room for improvements in these works. 

\section{Methods}
We first present the design basis and the overview of our system. Then the sensor planning algorithms to realize the active stereo vision are stressed. Finally, the dynamic obstacles modeling and the flight path planning approaches are described.

\subsection{System Design}
The system aims to enhance the observation performance of a single stereo camera to realize safe and rapid flight in dynamic environments.
Consider the attitude control model of a regular quadrotor \cite{QuadModel}. Let $\phi$, $\theta$, and $\psi$ denote roll, pitch, and yaw angle. The state vector of attitude is
\begin{equation}
\boldsymbol{x} = \left[ \phi \quad \dot{\phi} \quad \theta \quad \dot{\theta} \quad \psi \quad \dot{\psi} \right]^T
\label{eq:X}
\end{equation}

Then the simplified state space equations are usually expressed as:
\begin{equation}
\boldsymbol{\dot{x}} = \left[
    \begin{array}{c}
    \dot{\phi} \vspace{1ex} \\
    \dot{\theta}\dot{\psi}a_1 + \dot{\theta}a_2 \Omega_r + b_1 \Gamma_2 \vspace{1ex}\\
    \dot{\theta} \vspace{1ex}\\
    \dot{\phi}\dot{\psi}a_3 - \dot{\phi}a_4 \Omega_r + b_2 \Gamma_3 \vspace{1ex}\\
    \dot{\psi} \vspace{1ex}\\
    \dot{\theta}\dot{\phi}a_5 + b_3 \Gamma_4
    \end{array}
\right]
\label{eq:dotX}
\end{equation}
where $a_i$ and $b_i$ are the coefficients related to the quadrotor's size and weight, $\Gamma_i$ is the commonly used input force or torque, and $\Omega_r$ is the residual propeller speed. In particular, $\Gamma_4$ is provided by the torque rather than the lift force from the propellers, $a_5=(I_{xx}-I_{yy})/I_{zz}$ and $b_3=1/I_{zz}$, in which $I$ represents the moment of inertia.
For a quadrotor with symmetrical structure, $I_{xx} \approx I_{yy}$, which means $a_5 \approx 0$.
When $\dot{\psi}=0$, the system becomes a linear system that is prone to better control performance. 
\hl{In fact, controlling $\dot{\psi}$ during the flight would result in non-neglectable control error \cite{NonlinearControl} and is given with the lowest priority in \cite{YawControlNotGood} \cite{ThrustMixing}.}

Therefore, we prefer adding a ``head'' with additional DOF to realize active vision rather than swivel the yaw angle $\dot{\psi}$ of the quadrotor.
Define the angle of the head as $\xi$. It is worth noting that changes on $\xi$ would cause variations on $I_{xx}$ and $I_{yy}$, which makes the system time-varying and hard to be controlled. Thus we prefer a head that is very light and small compared to the whole quadrotor so that the variations on $I_{xx}$ and $I_{yy}$ can be neglected. One more advantage of a light and small head is that its moment of inertia $I^{h}_{zz}$ is small. Thus when the head is rotating, it requires less $\Gamma_4$ to provide reactive torque, which can be derived from:
\begin{equation}
\ddot{\xi}I^{h}_{zz} = \ddot{\psi}I_{zz} = \Gamma_4 + \dot{\theta} \dot{\phi} (I_{xx}-I_{yy}) \approx \Gamma_4
\label{eq:moment}
\end{equation}

Hence a stereo camera that takes only $3.3\%$ weight of the whole flying robot is applied to act as the rotatable head, and the other hardware components are fixed. Through numerical estimation in CAE software, $I^h_{zz}$ takes only 0.1$\%$ of $I_{zz}$ in our flying robot while the variations on $I_{xx}$ and $I_{yy}$ are less than $2\%$.
A regular flight controller running PID algorithms is enough for proper performance in this case. A servo motor with a high-precision encoder is connected under the stereo camera to form an active vision structure with one rotational DOF. Then the structure is placed on the top center of the quadrotor to have a clear view field, as is illustrated in Fig. \ref{MAV}(a).

\begin{figure}
\centerline{\psfig{figure=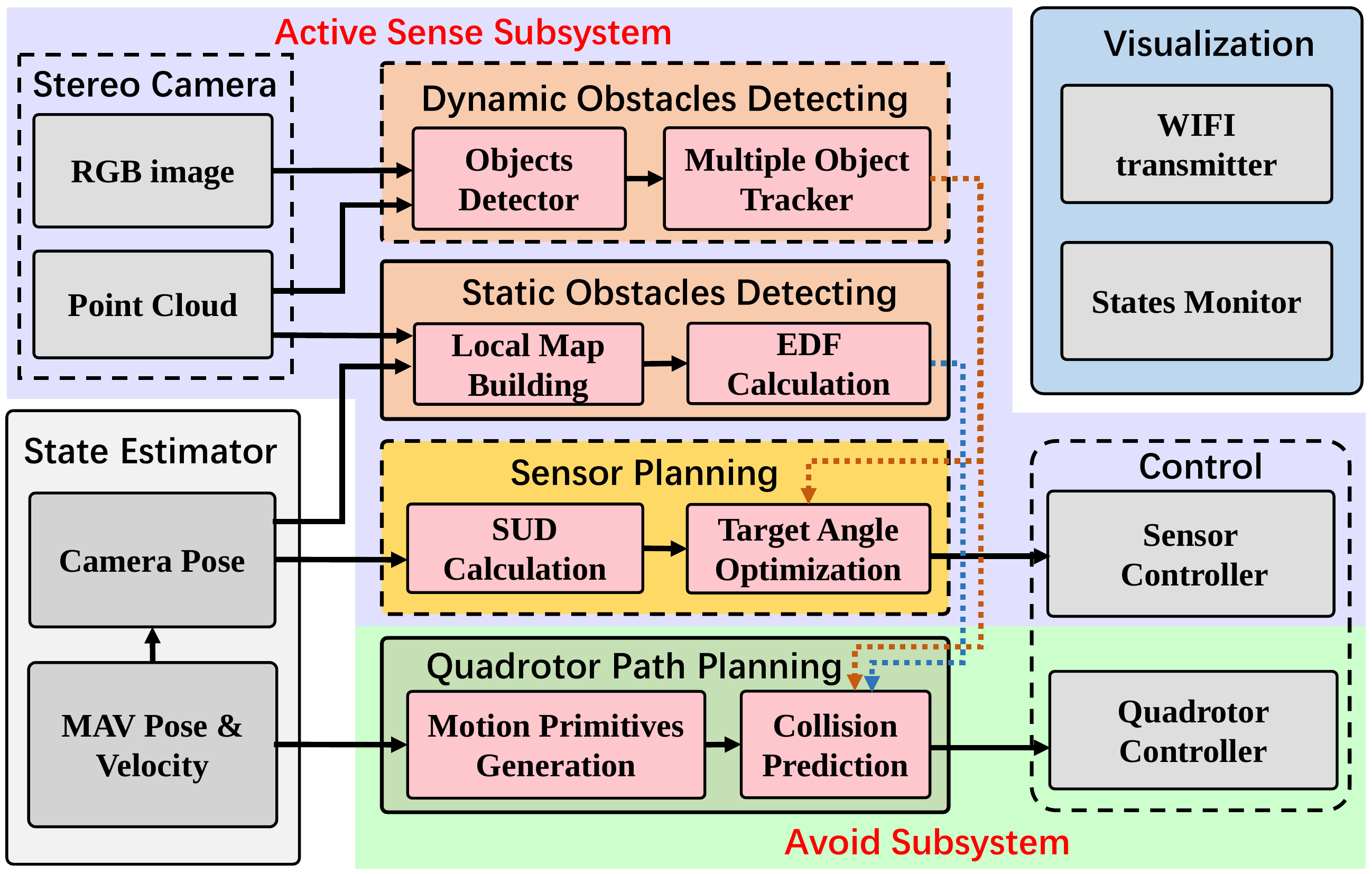,width=3.4in}}
\caption{An overview of our ASAA System.}
\label{ASAA}
\end{figure}

With this rotatable camera as the head, we designed a complete ASAA system, as Fig. \ref{ASAA} shows.
The main system could be divided into two subsystems: the active sense subsystem and the avoid subsystem. The former subsystem comprises two detectors designed respectively for dynamic obstacles and static obstacles, and particularly, planner and controller for the camera rotation. The latter subsystem consists of the path planner and the controller of the quadrotor. Both controllers for the sensor rotation and the quadrotor flight are simple PID controllers. The rest components are specified in the following subsections.

\subsection{Sensor Planning}
\begin{figure}
\centerline{\psfig{figure=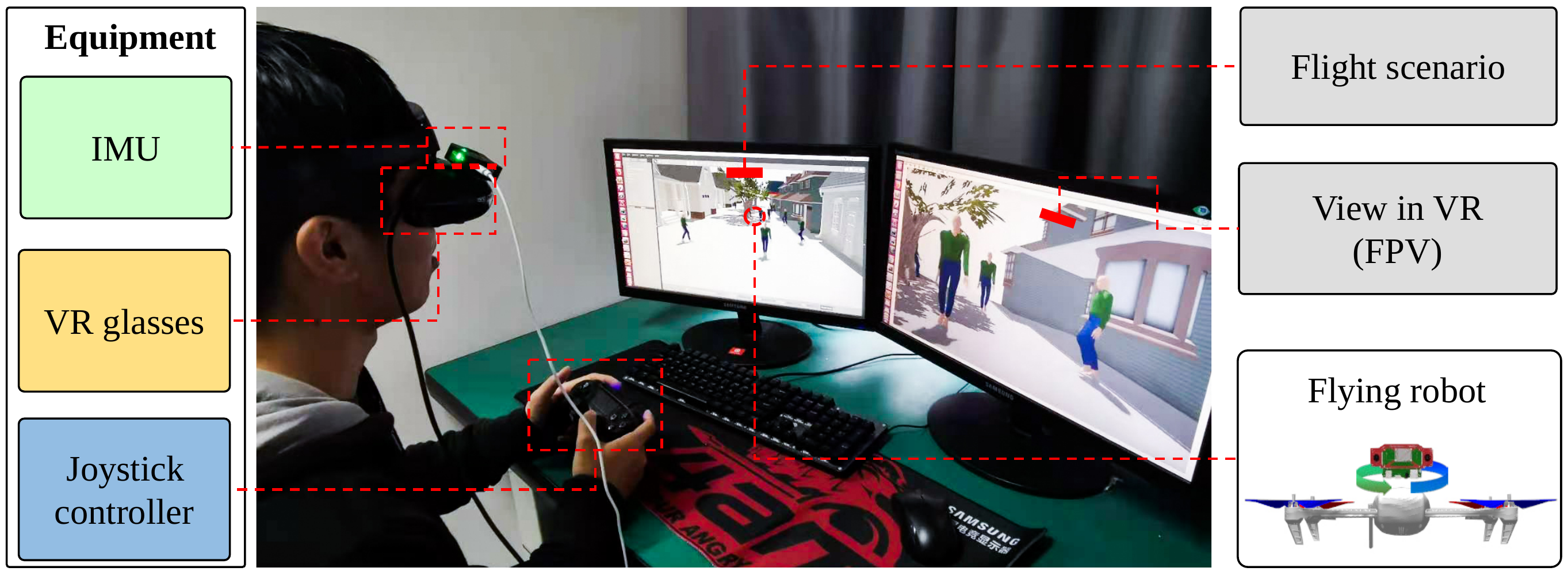, width=3.4in}}
\caption{\hl{Human expert controlling a flying robot with a rotatable camera in simulation through virtual reality glasses. The rotation of the camera follows the expert's head while the flight is controlled by a joystick controller.}}
\label{human_operator}
\end{figure}

Sensor planning in our system is to plan the stereo camera's rotation angle for a FOV that benefits flight safety. Since the environment is unknown and potential dynamic obstacles exist, this planning is a complex high-dimension problem of which a complete or optimal solution can hardly be found.
\hl{ We studied this sensor planning problem by referring to the experience of human operators. As Fig. \ref{human_operator} illustrates, an immersive simulation environment was built to train the human operators to control the flying robot with the ASAA system. The specification of the environment can be found in Subsection} \uppercase\expandafter{\romannumeral5}.A. \hl{The operators saw the first-person view of the camera through VR glasses. The FOV was 80 degrees, and the visible distance was only 10 $m$, which was to simulate the range of a real stereo camera. An IMU was mounted on the VR glasses to measure the operator's head rotation and control the camera rotation synchronously. The flight motion was controlled by a joystick controller.}

\hl{Since the environment was complex and the FOV was very limited, the operators had to rotate the camera to different directions to have a better observation. We then tried to find out the intentions of making these rotations.
When the operators could finish the task expertly, we asked them about their observation directions of interest during the flight. Ten operators participated, and four common concerned directions were summarized.
The first is $\boldsymbol{D}_g$, which is the direction of the goal position. The second is $\boldsymbol{D}_v$, which is the real current flight direction. $\boldsymbol{D}_g$ and $\boldsymbol{D}_v$ can be very different if an abrupt goal direction change occurs. The third is the directions where dynamic obstacles exist. More observation in these directions can help predict the path of dynamic obstacles. The last direction of interest is the directions that haven't been seen recently, which should also be concerned because unknown dynamic obstacles can appear anywhere and cause collisions. According to the operators' experience, a multi-objective sensor planning function is abstracted. Let $f_1$ to $f_4$ denote the cost functions of observing the above four directions, respectively, and $f_5$ denote an objective to resist large direction changes. $f_5$ is added to avoid frequent shaking that would cause motion blur in real applications.} The planner is described as:
\begin{equation}
\mathop{\arg}\min_{\xi_t} \ F(\xi_t)=\left[ f_1(\xi_t), f_2(\xi_t), ... , f_5(\xi_t)\right]^T
\label{eq:multiObjective}
\end{equation}

Since our camera has only one rotational DOF along the $z$ axis, we mainly concern the directions in the $xy$ plane.
Suppose the horizontal view angle of the camera is $\theta_h$, and the projection angle of each direction $\boldsymbol{D}_*$ in $xy$ plane is $d_*$.
Then
\begin{equation}
f_1(\xi_t)=G(\xi_t-d_g)\cdot \left(1- U(d_g)\right)
\label{eq:f1}
\end{equation}
where $U(*)$ is the function to get the SUD of a specific direction, which is given in \uppercase\expandafter{\romannumeral3}.C. $G(*)$ describes penalty when a direction is not in the camera's FOV. Suppose the angle of a direction is $\bar{\theta}$. $G(\bar{\theta})$ is given by:
\begin{equation}
G(\bar{\theta}) = \left\{
    \begin{array}{c}
    0, \quad if \ \bar{\theta} \in [-\frac{\theta_h}{2}, \frac{\theta_h}{2}] \\
    \bar{\theta}^{2} - \frac{\theta_h^2}{4}, \quad otherwise
    \end{array}
\right.
\label{eq:Gx}
\end{equation}

A quadratic function is used in (\ref{eq:Gx}) to obtain a large penalty when $\bar{\theta}$ is significantly far away from the FOV. Let $v_h$ denote the current horizontal velocity of the flying robot. While the velocity increases, more attention should be given to $d_v$. Since the attention towards $d_v$ is crucial when $v_h$ is large, we define this tendency as quadratic growth. 
$f_2(\xi_t)$ can be then expressed as:
\begin{equation}
f_2(\xi_t) = v_h^2 \cdot G(\xi_t-d_v) \cdot \left(1- U(d_v)\right)
\label{eq:f2}
\end{equation}

$f_3(\xi_t)$ is related to the state estimation result of the dynamic obstacles. If the currently estimated relative position and relative velocity of the $j_{th}$ dynamic obstacle are $\boldsymbol{p}_{o}^j$ and $\boldsymbol{v}_{o}^j$ respectively, and $\boldsymbol{p}_{o}^j$ lies in the direction $d_{o}^j$, then $f_3(\xi_t)$ can be described as:

\begin{equation}
f_3(\xi_t) = \sum_{j=1}^{N}  \frac{\beta |\boldsymbol{v}_{o}^j| }{|\boldsymbol{p}_{o}^j|}  G(\xi_t - d_{o}^j)
\label{eq:f3}
\end{equation}
where $N$ is the number of dynamic obstacles in tracking. The obstacle with smaller $|\boldsymbol{p}_{o}^j|$ and larger $|\boldsymbol{v}_{o}^j|$ should be valued more. $\beta$ is a coefficient defined to adjust the relative importance of $|\boldsymbol{v}_{o}^j|$ towards $|\boldsymbol{d}_{o}^j|$.

The forth cost function concerning exploring the less-updated area is simply given by
\begin{equation}
f_4(\xi_t)=U(\xi_t).
\label{eq:f4}
\end{equation}

The last cost function is to resist large direction changes. Let $\xi_{t-1}$ denote the rotation angle planned last time. $f_5(\xi_t)$ is calculated by:
\begin{equation}
f_5(\xi_t) = (\xi_t-\xi_{t-1})^2
\label{eq:f5}
\end{equation}

Then an overall cost function can be given to simplify the problem.
\begin{equation}
F(\xi_t)= \sum_{i=1}^{5} \lambda_i f_i(\xi_t)
\label{eq:Fx}
\end{equation}
where $\lambda_i$ are the weighting coefficients.

To find the optimal solution with the simplest approach, the direction angle is discretized with resolution $\delta$ and the optimal solution $\xi_t^*$ is searched by enumeration. If the camera can rotate freely, the searching domain of $\xi_t$ is $\mathbb{S}=\left\{ \xi_0-\pi, \xi_0-\pi+\delta, \xi_0-\pi+2\delta, ... , \xi_0+\pi-\delta \right\}$, where $\xi_0$ is the real current angle. If the camera has a limited rotation range $[\xi_{min}, \xi_{max}]$ that satisfies $\xi_{min} < -\pi-\delta$ and $\xi_{max} > \pi+\delta$ (a margin larger than $\delta$ is required to avoid unnecessary large rotation), the searching domain is $\mathbb{S} = \left\{ -\pi-\delta, -\pi, -\pi+\delta, ... , \pi+\delta \right\}$. In our case, the rotation range is limited by the cable on the camera thus the latter domain is adopted. 

The behavior of the result camera rotation is comprehensive. Different states of the flying robot and the environment could lead to different behaviors.
When the flying robot is hovering with no goal position received, $f_1$ and $f_2$ are zero for all angles. If no dynamic obstacle is detected, the camera rotates in a range of 360 degrees to scan the whole neighborhood area due to the effect of $f_4$ and $f_5$. When a goal position is received, but the flying robot is still hovering because the flight is not allowed or the robot is trapped in a very dense area, $f_1$, $f_4$, and $f_5$ take effect, and the camera rotates in a relatively small range around the goal direction to search for a valid flight path.
If dynamic obstacles appear, the camera will pay more attention to them due to $f_3$.
When the flying robot is moving, $f_2$ takes effect, and the behavior of the camera is a comprehensive result of all the five optimization components.

\subsection{SUD Estimation}
Our SUD is used to assess the fully updated probability of each direction in $\mathbb{S}$. A fully updated direction suggests all the obstacles along this direction are observed lately. \hl{ A significant factor that affects the SUD is the rotation of the stereo camera. Generally, the updated probability of the directions inside of the FOV should increase while the probability of the rest directions should decrease. Besides, the motion of the flying robot also affects this probability. Let Vector $\Delta\boldsymbol{p}$ denote the flying robot's position displacement during one calculation period.
Since the unknown area comes along the direction of $\Delta\boldsymbol{p}$, the fully updated probability near this direction should decrease accordingly. Let $y_t$ denote the current measurement related to the rotation of the camera and the motion of the flying robot. The log-odds increment of the fully updated probability of $d_i \in \mathbb{S}$ regarding measurement $y_{t}$ can be expressed as:}
\begin{equation}
L(d_i|y_t)= -\varepsilon_1 \Delta  \boldsymbol{p} \cdot \boldsymbol{\bar{d}}_i L_h^{-1} - \varepsilon_2 | \Delta \boldsymbol{p} \cdot \boldsymbol{k}| + l(d_i)
\label{eq:L}
\end{equation}
\hl{where $\boldsymbol{\bar{d}}_i$ represents the unit direction vector of $d_i$, $L_h$ describes the valid depth measurement distance of the stereo camera and $\boldsymbol{k}$ represents a unit vector along the $z$ axis.
$\varepsilon_1$ and $\varepsilon_2$ are two weighting coefficients.
The first item and the second item in Eq. (\ref{eq:L}) give the probability decrease or increase caused by horizontal motion and vertical motion, respectively, via an exponential model \cite{OctoMap}. For the reason that the camera swivels in the $xy$ plane in body frame, we take $\varepsilon_2 > \varepsilon_1 = 1$ in practice.
$l(d_i)$ describes the log-odds increase of the probability when direction $d_i$ is in the camera view and is given by:}
\begin{equation}
l(d_i)= \left\{
    \begin{array}{l}
    l_{hit}, \quad if \ \vert d_i-\xi_0 \vert \leq \frac{\theta_h}{2} \\
    l_{miss}, \quad otherwise
    \end{array}
\right.
\label{eq:hit}
\end{equation}
\hl{with $l_{hit} > 0$ and $l_{miss} < 0$.}

\hl{Suppose the log-odds probability of one direction $d_i \in \mathbb{S}$ to be fully updated at a discrete time $t$ is $L(d_i|y_{1:t})$. Given the previous probability $L(d_i|y_{1:t-1})$ (initialized with zero), the update formula regarding current measurement $y_{t}$ can be expressed as:}
\begin{equation}
L(d_i|y_{1:t})=max(min(L(d_i|y_{1:t-1})+L(d_i|y_t) , l_{max}), l_{min})
\label{eq:update}
\end{equation}
\hl{where $l_{min}=0$ and $l_{max}=1$ describe the lower and upper bound. For the purpose of computational efficiency, $L(d_i|y_{1:t})$ of each direction is stored in a one-dimension buffer and Function $U(*)$ in Eq. (\ref{eq:f1}) and Eq. (\ref{eq:f2}) is to query the buffer to get the log-odds probability of one direction, which means $U(d_i) = L(d_i|y_{1:t})$.}


\subsection{Dynamic Obstacle Modeling}
To model and predict the trajectories of dynamic obstacles, a multiple object tracker (MOT) is adopted. Since all our computation is conducted in an onboard computer with quite limited computing power, a very efficient MOT is required. In addition, multiple dynamic obstacles could be around the flying robot in different directions, and some could only be seen occasionally due to the limited FOV and the camera motion, which leads to the occlusion problem \cite{MOTReview}.

Here we modified SORT \cite{SORT} to adapt to our requirements. SORT is a very light algorithm that first detects objects in the current image and then uses Kalman Filter (KF) \cite{kalmanFilter} and Hungarian algorithm for tracking and data association.
\hl{ The features estimated in the KF of SORT are the pixel position and the pixel velocity. However, this pixel-level estimation approach is heavily influenced by the occlusion problem caused by camera motion. Thus we instead estimate the global position and the global velocity of the objects in the world coordinate, which are independent of the camera motion and can be associated after occlusions. Specifically, the objects are first detected in the RGB image by the tiny version of YOLO V3 \cite{Yolov3}. Each object's depth is estimated by averaging the depth values around the center point of the bounding box in an aligned depth image. Then the position of the object in the camera coordinate is calculated by the pinhole model, and the position in the world coordinate is acquired through coordinate transformations. The shape of an object is considered as a cylinder. The height and the diameter of the cylinder can be similarly estimated by applying the pinhole model to the bounding box's vertexes.}


The motion model of each dynamic obstacle in the KF is defined as a linear constant velocity model with independent motion on each axis. Take the motion on the $x$ axis as an example. The acceleration of the $j^{th}$ obstacle is $a^j_{x} \sim N(0, {\sigma^j_{x}}^2)$, where ${\sigma^j_{x}}^2$ is the acceleration variance and is initialized according to the label of the obstacle.
\hl{Suppose the currently estimated state vector of the obstacle is $\boldsymbol{s}_{0}^j =\left[x_0^j,  \dot{x}_0^j\right]$, where the first element denotes position and the second element denotes velocity. The state vector $\boldsymbol{s}_{t}^j$ at a future time $t$ is modeled as:} 
\begin{equation}
\boldsymbol{s}_{t}^j =
\left[
\begin{array}{c}
  x_{t}^j \\
  \dot{x}_{t}^j
\end{array}
\right]
=
\left[
\begin{array}{cc}
  1 & t \\
  0 & 1
\end{array}
\right]
\left[
\begin{array}{c}
  x_{0}^j \\
  \dot{x}_{0}^j
\end{array}
\right]
+
\left[
\begin{array}{c}
  \frac{1}{2} a_{x}^j  t^2 \\
  a_{x}^j  t
\end{array}
\right]
\label{eq:prediction}
\end{equation}
The state distribution center is
\begin{equation}
\hat{\boldsymbol{s}}^j_t=\left[ x_{0}^j + \dot{x}_{0}^j t, \ \dot{x}_{0}^j \right]^\mathrm{T}
\label{eq:distribution_center}
\end{equation}
while the covariance matrix is calculated by
\begin{equation}
\begin{split}
   \emph{Var} \left( \boldsymbol{s}_{t}^j \right) & =E\left[ (\boldsymbol{s}_{t}^j-\hat{\boldsymbol{s}}^j_t)(\boldsymbol{s}_t-\hat{\boldsymbol{s}}^j_t)^\mathrm{T} \right] \\
     & = {\sigma^{j}_x}^2 \left[
     \begin{array}{cc}
       0.25 t^4 & 0.5 t^3 \\
       0.5 t^3 & t^2
     \end{array}
     \right]
\end{split}
\label{eq:covariance}
\end{equation}

\subsection{Flight Path Planning}
\hl{The flight path planning is running in real-time and is composed of two parts, which are motion primitives generating and collision checking. The state-space sampling method described in \cite{AggressiveSample} is adopted to generate motion primitives. Firstly, some temporal goal position candidates are uniformly sampled without the consideration of obstacles. Suppose $\boldsymbol{p}$ is one of the candidates.}
Since our flying robot is equipped with active stereo vision, the sampling region for $\boldsymbol{p}$ is not limited in FOV. Let $\boldsymbol{p}_{goal}$ denote final goal position and $\boldsymbol{p}_{rob}$ denote the current position of the flying robot.
Define two direction vectors:
\begin{equation}
\boldsymbol{l}_1 = \boldsymbol{p}_{goal}-\boldsymbol{p}_{rob}, \ \boldsymbol{l}_2 = \boldsymbol{p}-\boldsymbol{p}_{rob}
\label{smaple region2}
\end{equation}
The sampling region is
\begin{equation}
\mathbb{S}_p = \left\{\forall \boldsymbol{p} \in \mathbb{S}_p \big| \left< \boldsymbol{l}_1, \boldsymbol{l}_2 \right> < \theta_{val},
|\boldsymbol{l}_2| \leq  l_{vis} \right\}
\label{eq:smaple region}
\end{equation}
where $l_{vis}$ is the depth visible distance, and \hl{$\theta_{val} \in [0, \pi]$ is a parameter that controls the valid angle range of $\mathbb{S}_p$.} If $\theta_{val} > \frac{\pi}{2}$, temporal goal position candidates that lead to the opposite direction of $\boldsymbol{l}_1$ might be sampled. Selecting these candidates would temporarily lead to a path away from the final goal but is conducive to flight safety in the environment with dynamic obstacles. For instance, when a dynamic obstacle comes closely in a sudden and no collision-free motion primitive to the final goal direction can be found, a decent strategy is to fly to a posterolateral position temporarily to avoid the collision.
If $\theta_{val} = \pi$, $\mathbb{S}_p$ represents a sphere with a radius of $l_{vis}$, which suggests a real-time omnidirectional observation can be acquired. For the reason that the rotation speed of our camera is limited, the direction opposites to $\boldsymbol{l}_1$ is usually unable to be observed immediately. Thus $\theta_{val} = \frac{2}{3} \pi$ is taken.

\hl{Then all the temporal goal position candidates are sorted by the cost functions described in \cite{AggressiveSample} and the motion primitive from the current position to each candidate can be generated \cite{MullerStatetoState}.}
Let $\boldsymbol{p}_m^i(t), t\in[0, T]$ denote the position setpoint on the $i_{th}$ generated motion primitive at a future time $t$, where $T$ is the estimated flight time in motion primitive generation \cite{MullerStatetoState} and $\boldsymbol{p}_m^i(0)=\boldsymbol{p}_{rob}$. Our collision checking approach is described as follows.

The collision checking considers both static obstacles and dynamic obstacles. The static obstacles are represented by an egocentric local occupancy map \cite{Ringbuffer}. The voxels in the position of dynamic obstacles are set free directly to eliminate the noise caused by them. Then the Euclidean distance field (EDF), which describes the distance from each free voxel to the nearest obstacle, is used to check the closest distance to static obstacles.

\hl{Since the predicted position of a dynamic obstacle from the KF obeys Gaussian distribution, the Mahalanobis distance \cite{MahalanobisDistance} is adopted to calculate the distance from $\boldsymbol{p}_m^i(t)$ to the dynamic obstacle.
Compared to the Euclidean distance, Mahalanobis distance can take state estimation uncertainty into account \cite{SORT2}.}
\hl{Suppose the predicted position distribution center of the $j_{th}$ dynamic obstacle at time $t$ is $\boldsymbol{p}_{o}^j(t)$, which is calculated with Eq. (\ref{eq:distribution_center}). According to the cylinder shape model in Subsection} \uppercase\expandafter{\romannumeral3}.D, \hl{$\boldsymbol{p}_{o}^j(t)$ indicates the position of the cylinder's center. To evaluate the distance between the flying robot and the dynamic obstacle, the radius and the height of the cylinder model should be concerned.
Let $r_{o}^j$ and $h_{o}^j$ denote the cylinder's radius and height, respectively.
If the height of $\boldsymbol{p}_m^i(t)$ is smaller than $h_{o}^j$, the height of $\boldsymbol{p}_m^i(t)$ is further assigned to $\boldsymbol{p}_{o}^j(t)$. Otherwise, the height of $\boldsymbol{p}_{o}^j(t)$ is set to be $h_{o}^j$.
Then the vector from $\boldsymbol{p}_m^i(t)$ to its closest point on the dynamic obstacle is}
\begin{equation}
\Delta \boldsymbol{p}_{i}^j(t) = \left(\boldsymbol{p}_m^i(t) - \boldsymbol{p}_o^j(t) \right) \left( 1 - \frac{r_{o}^j}{|\boldsymbol{p}_m^i(t) - \boldsymbol{p}_o^j(t)|}  \right)
\label{eq:Corrected_delt_p}
\end{equation}
\hl{The second term of this multiplication is added to fulfill the correction along the radial direction.}
Then the squared form of the Mahalanobis distance is:

\begin{equation}
{d_M}_i^j(t) = \Delta \boldsymbol{p}_{i}^j(t)^{T} \ \Sigma_j^{-1}(t) \ \Delta \boldsymbol{p}_{i}^j(t)
\label{eq:MDistance}
\end{equation}
According to Eq. (\ref{eq:covariance}),
\begin{equation}
\Sigma_j(t) = 0.25 t^4 \left[
\begin{array}{ccc}
  {\sigma_{x}^j}^2 & 0 & 0 \\
  0 & {\sigma_{y}^j}^2 & 0 \\
  0 & 0 & {\sigma_{z}^j}^2
\end{array}
\right]
\label{eq:Sigma}
\end{equation}

Assume that the distance of $\boldsymbol{p}_m^i(t)$ to its closest obstacle, among static or dynamic obstacles separately, is $D(\boldsymbol{p}_m^i(t))$. The $i_{th}$ motion primitive is safe if $\forall t \in (0,T]$, the following condition is true for both static and dynamic obstacles:
\begin{equation}
D(\boldsymbol{p}_m^i(t)) > D_{min} \vee \dot{D}(\boldsymbol{p}_m^i(t)) > 0
\label{eq:collision free}
\end{equation}
\hl{where $D_{min}$ is the distance threshold. The $D_{min}$ for dynamic obstacles and the $D_{min}$ for static obstacles are different values tuned by experience. The former condition in (\ref{eq:collision free}) describes that the setpoint too close to obstacles is unsafe. The latter condition is considered because the dynamic obstacles might unexpectedly come to a very close distance (less than $D_{min}$) to the flying robot, and the flying robot might also intrude to the area close to the obstacles due to control error or external disturbance.
In these cases, the former condition can not be satisfied no matter which direction the primitive leads.
However, if $\dot{D}(\boldsymbol{p}_m^i(t)) > 0$ is satisfied, it suggests that the primitive is leading to a position far from the obstacles, which should also be treated as a safe path.}
Take the setpoints discretely with time interval $\Delta t$ and let $\boldsymbol{p}_{m,k}^i$ denote the $k_{th}$ ($ k\in\{1,2,...,\frac{T}{\Delta t}\}$) position setpoint on the $i_{th}$ motion primitive. Eq. \ref{eq:collision free} turns to $\forall k\in\{1,2,...,\frac{T}{\Delta t}\}$,
\begin{equation}
D(\boldsymbol{p}_{m,k}^i) > D_{min} \vee D(\boldsymbol{p}_{m,k}^i) \geq D(\boldsymbol{p}_{m,k-1}^i)
\label{eq:collision free_discrete}
\end{equation}
Equal to is taken in the latter condition because the distance resolution in EDF is limited. Two adjacent voxels in the local map might have the same distance value.

\section{Implementation}
Two essential details during the implementation are emphasized in this section. The first is to reduce the depth estimation error of the stereo camera, while the second is to synchronize the data from different sources. The following presents our strategies for these details.

\subsection{Depth Estimation}
Rotating the camera provides an agile way for observation while motion blur would occur, and the depth estimation would be inaccurate. Let $\boldsymbol{v}_{cam }^j$ denote the relative velocity of an obstacle $j$ to the camera, and $d_{cam}^j$ denote the distance between the camera and the obstacle. $\boldsymbol{v}_{cam}^j$ is derived by
\begin{equation}
\boldsymbol{v}_{cam}^j = d_{cam}^j (\dot{\xi}+\dot{\psi}) \boldsymbol{k}_{cam} + \boldsymbol{v}_{rob}-\boldsymbol{v}^j_o
\label{eq:relative velocity}
\end{equation}
where $\boldsymbol{k}_{cam}$ is a unit vector perpendicular to the camera direction in the $xy$ plane. $\dot{\psi}=0$ is taken in the experiments because the rotation for observation is achieved by $\dot{\xi}$. $|\boldsymbol{v}_{cam}^j|$ can be very large if $\dot{\xi}$ is large, which would result \hl{in} terrible motion blur phenomenon. The measurement for the depth of obstacles would be inaccurate.

\begin{figure}
\centerline{\psfig{figure=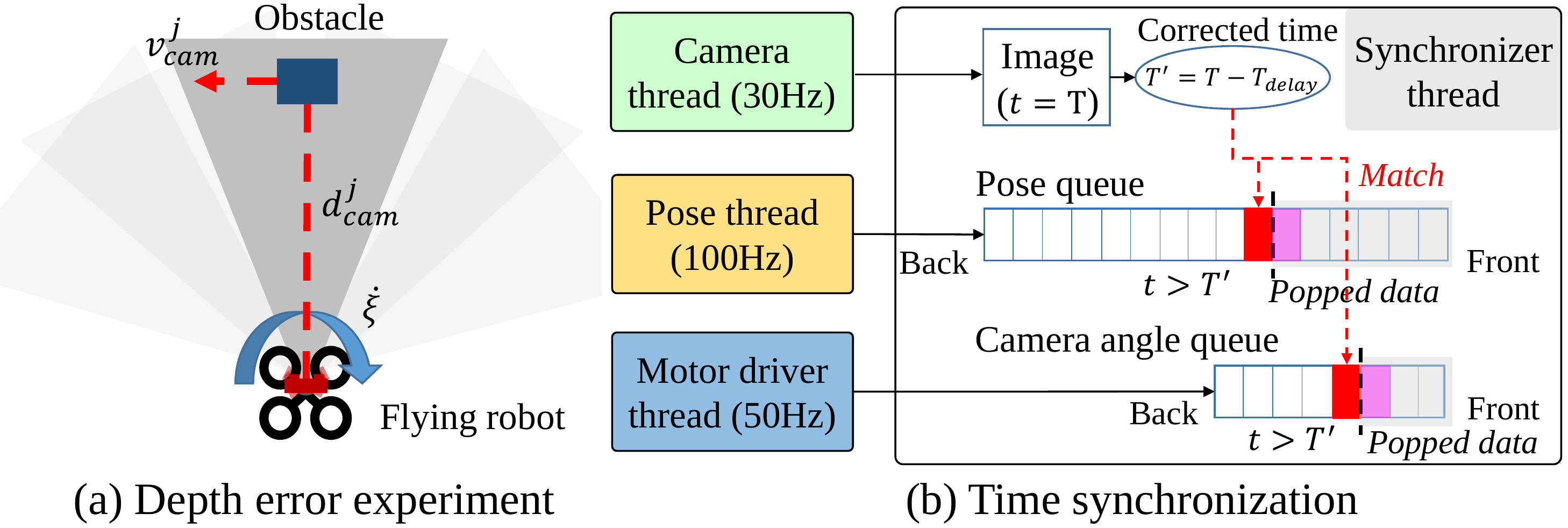,width=3.4in}}
\caption{Illustration of our experiment to estimate depth error caused by motion blur and the working mechanism of the synchronizer.}
\label{synchronizer}
\end{figure}

\begin{figure}
\centerline{\psfig{figure=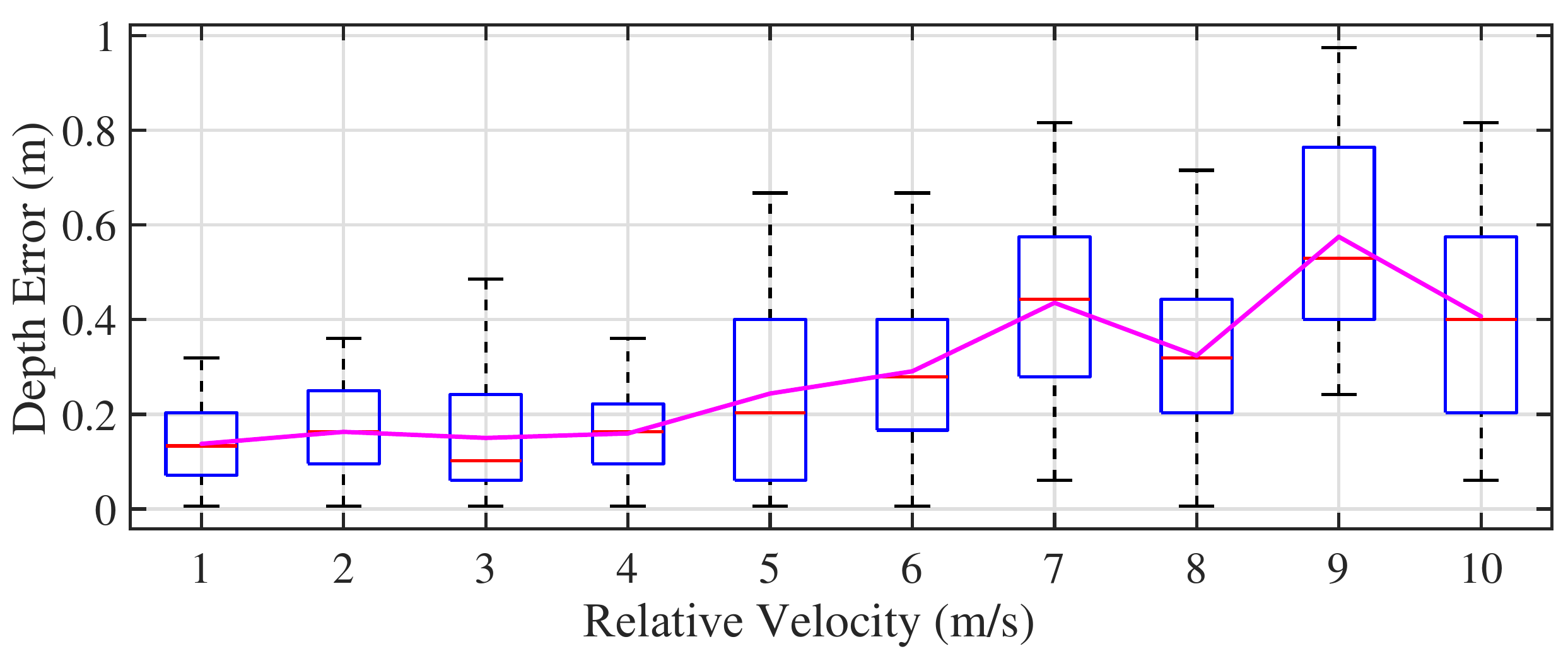,width=3.2in, height=1.3in}}
\caption{Depth estimation error at different relative velocities.}
\label{error_curve}
\end{figure}

Deblurring methods can alleviate motion blur, but they consume computation resources and cause more latency. Therefore, we instead tune the camera parameters, such as exposure time and brightness, to achieve a faster shutter and \hl{set a limitation $\dot{\xi}_{max}$ for the maximum rotation speed to keep the error caused by motion blur within an affordable level. The value of $\dot{\xi}_{max}$ is investigated through an experiment shown in Fig. \ref{synchronizer}(a). We set different $\dot{\xi}$ and measure the depth error $e_d$ of an obstacle in the depth image when the obstacle is in front of the camera. Both the flying robot and the obstacle are static during the test. Thus, the relative velocity is $\boldsymbol{v}_{cam}^j=d_{cam}^j\dot{\xi}$, where $d_{cam}^j$ is 3.0 m. The experiment result with a Realsense D435 camera is shown by a box plot with average value curve in Fig. \ref{error_curve}. In practice, the maximum permissible depth error when $d_{cam}^j=3.0$ m is set to be 0.2 m, which means $|\boldsymbol{v}_{cam}^j | \leq 4.5$ m/s and $\dot{\xi}_{max}=1.5$ rad/s.} Considering the linear motion of the flying robot and the dynamic obstacles, we set $\dot{\xi}_{max}=1.2$ rad/s in real-world experiments.
%

\subsection{Time Synchronization}
Since the camera could rotate rapidly, images or point clouds collected between two adjacent frames could be quite different. Synchronizing the data collected from different sources is very important. The data include images and point clouds from the stereo camera, the body pose estimated by the flight controller, and the camera rotation angle from the servo motor, whose updates are fulfilled by independent threads with the rates of 30Hz, 100Hz, and 50Hz, respectively. \hl{Our synchronization approach is illustrated in Fig. \ref{synchronizer}(b).
Firstly, the time delay in the data collection procedure is considered.
The body pose estimation and the camera rotation angle estimation are fulfilled in a few milliseconds, and thus the delay is ignored.
However, collecting images or point clouds from the stereo camera have a nonnegligible delay.
We measured the average delay time $T_{delay}$ in the image collection process by facing the camera to a digital clock and comparing the captured time with the real clock time. A more accurate timestamp of the image can then be given by $T'=T-T_{delay}$.
To further synchronize the image with the robot pose and the camera angle, two queues are adopted to store the pose and the angle data.
The newly arrived data with timestamp is pushed back to the queue. And the data stored in the queue is popped until an element with a timestamp large than $T'$ is found.
Then the synchronized result is chosen from this element (red element) or the last popped element (violet element), depending on whose timestamp is closer to $T'$. The average time difference after the synchronization is below 10 ms. Furthermore, a more accurate synchronized result can be achieved by linear interpolation with the two colored elements in the queue.}


\begin{figure}
\centerline{\psfig{figure=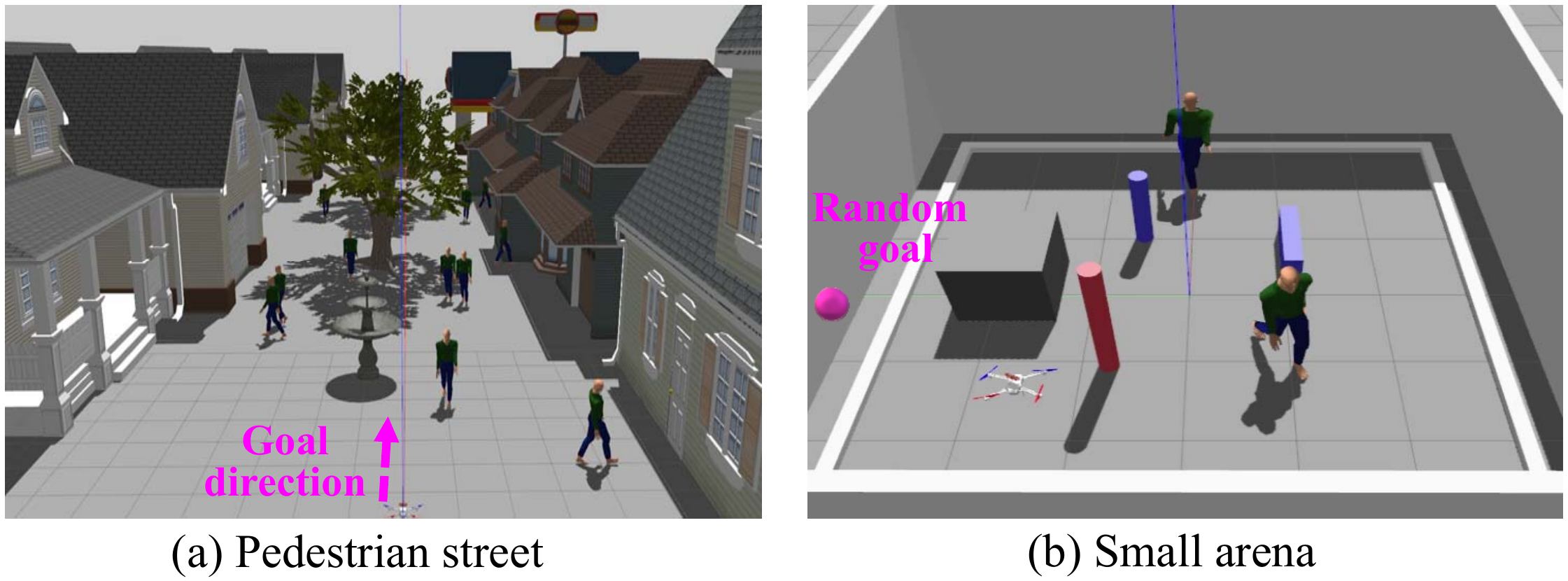, width=3.3in}}
\caption{Simulation tests scenarios built in Gazebo.}
\label{sim_scenarios}
\end{figure}

\section{Experiments}
\hl{Our system was tested in both simulation and real-world environments. In simulation tests, our system was compared with the traditional fixed-camera obstacle avoidance systems in two dynamic scenarios. Furthermore, real-world tests were conducted to demonstrate the effectiveness of this system in dealing with dynamic obstacles and abrupt goal direction changes.}

\subsection{Simulation Tests}
\begin{figure}
\centerline{\psfig{figure=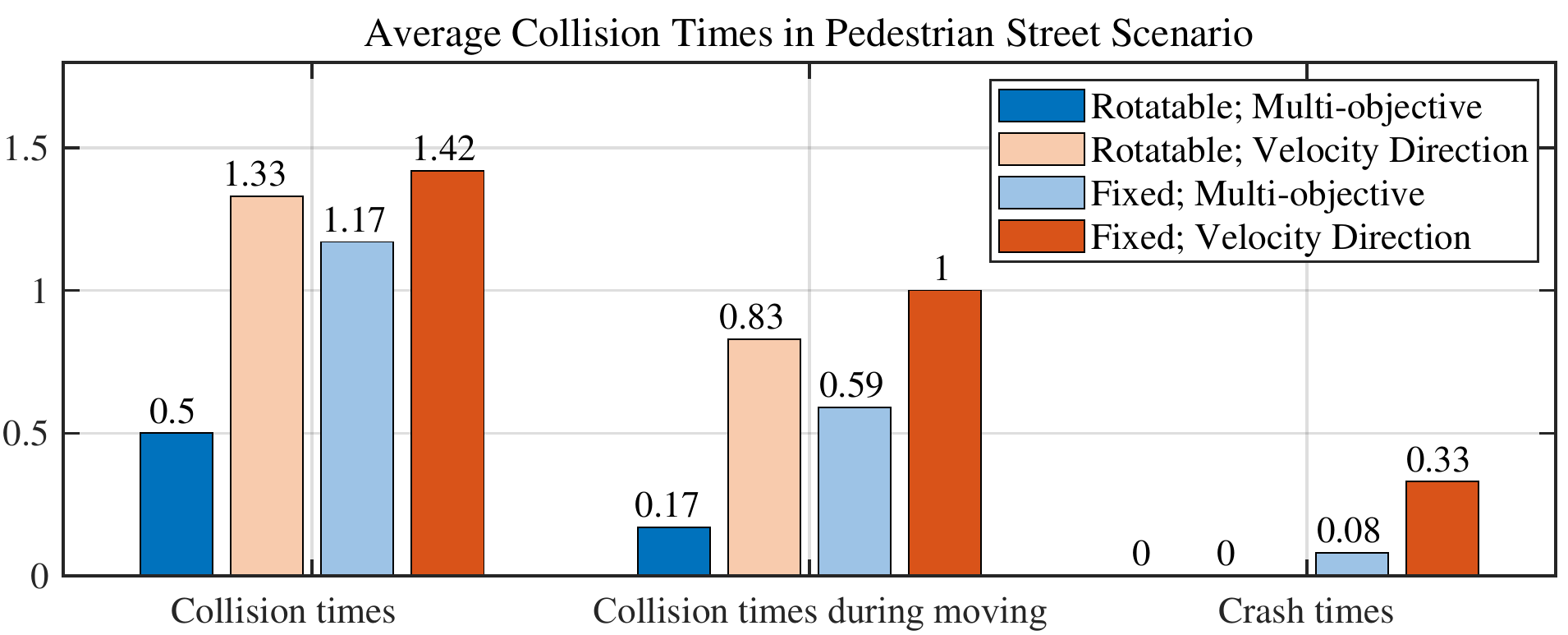, width=3.4in}}
\caption{\hl{Simulation test results in the pedestrian street scenario.}}
\label{pedestrian_street}
\end{figure}

\begin{figure}
\centerline{\psfig{figure=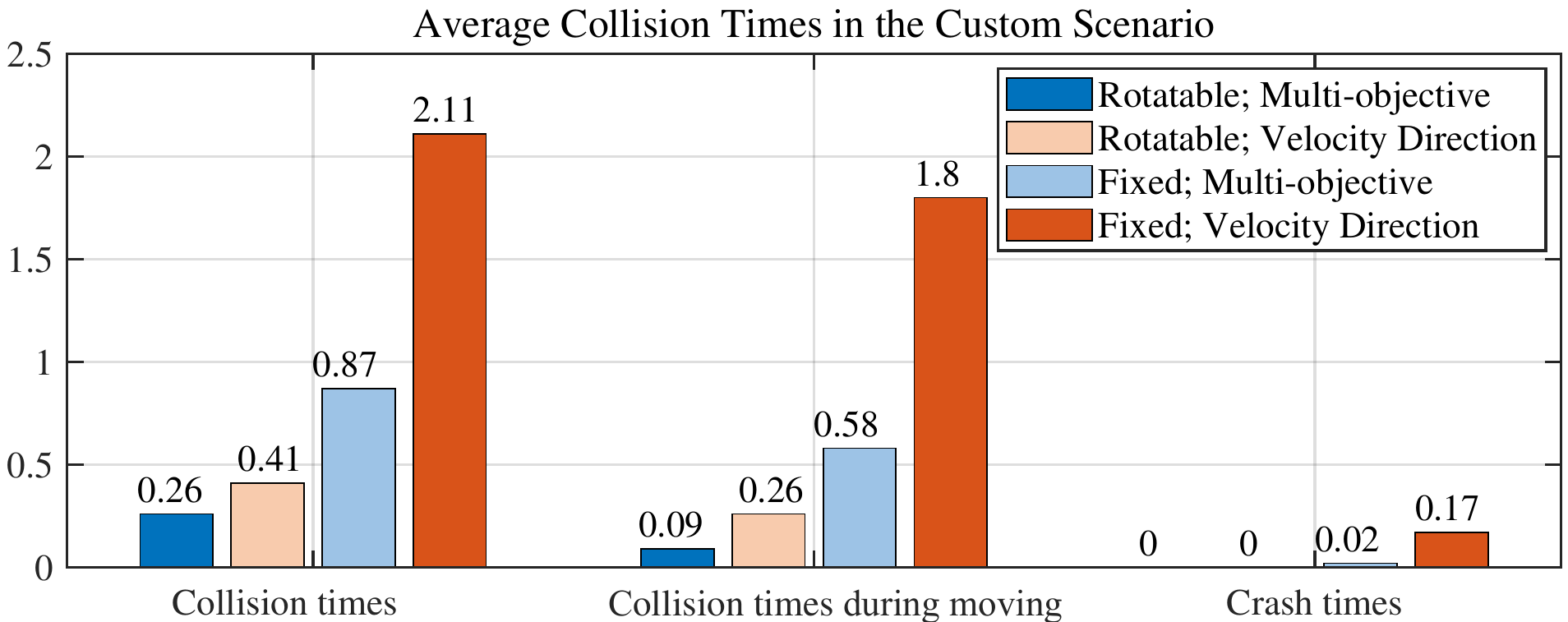, width=3.4in}}
\caption{\hl{Simulation test results in the small arena scenario.}}
\label{201world}
\end{figure}

\hl{Gazebo}$\footnote{http://gazebosim.org/}$ \hl{simulation environment and the RotorS \cite{Rotors} framework were used in the simulation tests. A VI-sensor with a continuous joint was mounted on the top of an IRIS quadrotor \cite{Rotors} to construct our ASAA system. We compared the performance of our system with three other paradigms, namely observing only the velocity direction with the rotatable camera, observing multi-objective planned direction by yaw rotation, and observing only the velocity direction by yaw rotation \cite{DAAdynamic}, \cite{SoftYawConstrain}. In the last two paradigms, the camera was fixed. All paradigms shared the same MOT and the same flight path planner.

Two testing scenarios with dynamic obstacles were built in Gazebo (Fig. \ref{sim_scenarios}). The first scenario was a 50-meter-long pedestrian street where many pedestrians walked around with different velocities, varying from 0.5 $m/s$ to 1.5 $m/s$. The goal of the flying robot was to reach the end of the street and avoid collisions along the way, which is a common task without the consideration of goal direction changes. Twelve tests were conducted for each paradigm in this scenario. The second scenario simulated a more challenging task, where the flying robot played the game of an owl trying to catch a rat (goal) in a small arena. The rat was so smart and agile that it would escape and show up at a random position on the other side whenever the flying robot approached it, which modeled the abrupt goal direction changes. The arena was a six-meter square with static obstacles and dynamic obstacles. For each paradigm, one hundred times of goal direction changes were tested. The safety threshold $D_{min}$ was 0.5 m (Euclidean distance) for static obstacles and 0.65 (Mahalanobis distance) for dynamic obstacles. Table \ref{Tabel:Coefficients} describes the coefficients for camera rotation planning in the test. The coefficients were tuned by trial and error.

Since the final purpose of designing the ASAA system is to realize better obstacle avoidance performance, we evaluated our paradigm and the other three paradigms by the average collision times in each test.
When the flying robot encountered a very complex situation with many pedestrians around, it was very likely that no feasible path could be found. If this situation occurred, a hover strategy was taken to wait for a feasible path. It should be noted that during the hover time, pedestrians might still collide with the flying robot because they were unable to sense the flying robot in the simulation. This collision should not be considered as the flying robot's failure, and thus the collision times when the robot was moving were further calculated. In addition, since our path planner plans from the current state of the quadrotor and replans in real-time, the flight could be recovered automatically after a slight collision in the simulation. When a severe collision happened, the flying robot crashed and the situation was counted separately. The results are shown in Fig. \ref{pedestrian_street} and Fig. \ref{201world}.}

\begin{figure}
\centerline{\psfig{figure=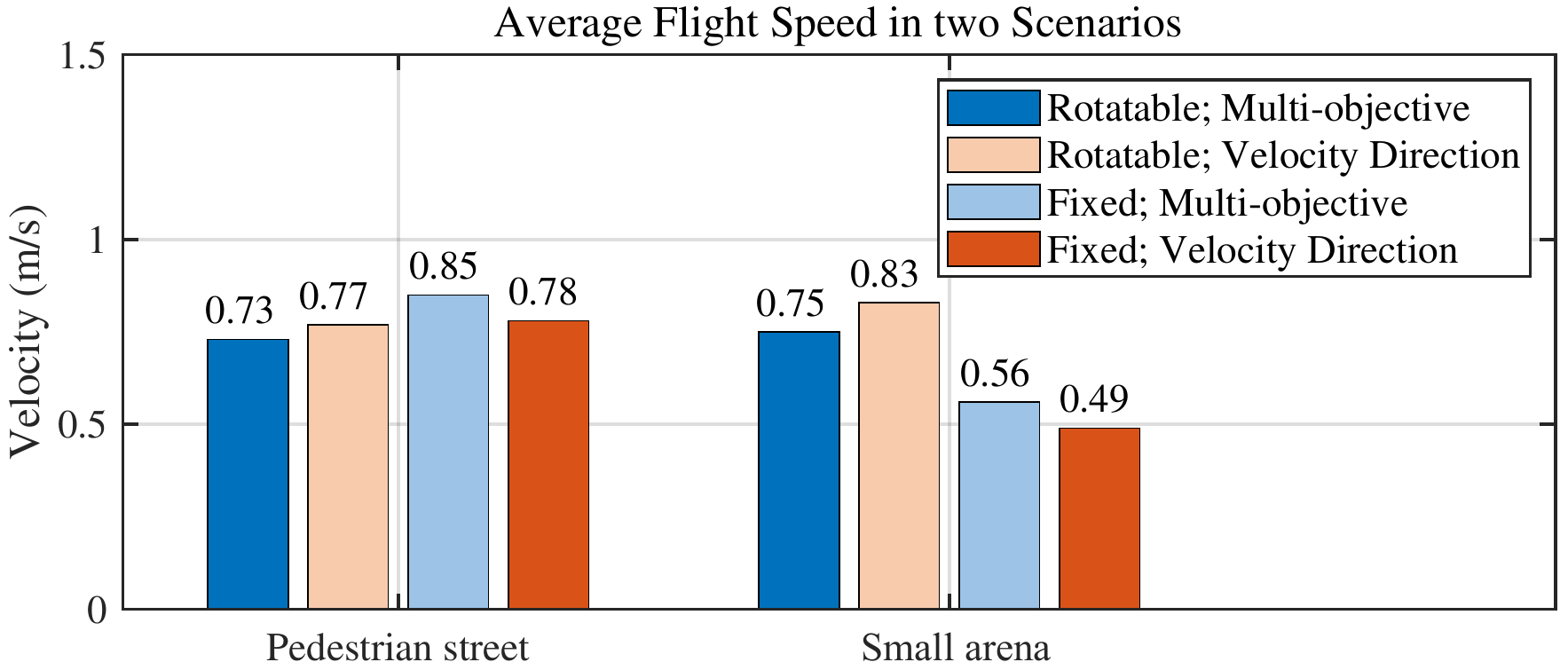, width=3.2in}}
\caption{\hl{Average flight speed in simulation tests.}}
\label{sim_velocity}
\end{figure}

\hl{In both simulation scenarios, our ASAA system performs the best results. The average collision times during moving are $83\%$ and $95\%$ smaller than those of the traditional paradigm that observes only the velocity direction with a fixed camera \cite{DAAdynamic}, \cite{SoftYawConstrain}, respectively.
Using a rotatable camera and observing the multi-objective planned direction can both improve the performance.
When abrupt goal direction changes exist (the small arena scenario), using a rotatable camera reduces the average collision times during moving by $85\%$ and observing the multi-objective planned direction reduces that by $67\%$.
When the camera is fixed, the flying robot has to rotate the yaw angle to observe different directions. The flight becomes unstable, and several crashes happened in both scenarios. In comparison, no crashes occurred when the camera is rotatable.
The average flight speed is further presented in Fig. \ref{sim_velocity}. In the pedestrian street scenario, the average flight speed of using a rotatable camera is slightly slower than that of using a fixed camera because more obstacle avoidance maneuvers were performed. In the small arena scenario with goal direction changes, using a rotatable camera is over $25\%$ faster because less time was spent on turning.}

\subsection{Real-world Tests}
A quadrotor with an axle base of 400 mm (Fig. \ref{MAV}) was adopted in the real-world tests. Realsense D435 with a horizontal view field of $72^\circ$ was applied to be the ``head''. And a servo motor with an encoder whose resolution is $0.088^\circ$ was utilized to be the ``neck''. A UP Core plus computing board running ROS was adopted as the onboard computer. The mounted CPU was Intel Atom x7 (4 Cores, 1.8 GHz).  A Vision Plus module with three Myriad X neural computing chips was also equipped to run the YOLO V3 (Tiny) detection model.
Our flight controller was Pixhawk running PX4 firmware. Pose estimation came from the Vicon motion capture system during the test.
We intended to test the real-time sense and agile obstacle avoidance ability of our system thus the length of the egocentric local map was set to be merely 6.4 m and $l_{vis}$ was set to be 1.5 m, which means the sensing range was merely about 3.2 m and the avoidance maneuver had to be taken within 1.5 m to the obstacles.\hl{ The safety thresholds were the same as those used in the simulation tests.}
\hl{The coefficients for camera rotation planning were tuned by trial and error in simulations.} The maximum velocity of the flying robot was set to be 1 m/s. The dynamic obstacles were several ground vehicles loaded with foam pillars (yellow top with Letter ``R'' in our figures), which moved at a maximum speed of 0.5 m/s. The resolution $\delta$ to update the SUD was $\frac{\pi}{8}$.

\begin{table}[]
\caption{Coefficient Values in Experiments}
\centering
\begin{tabular}{llllllllll}
\hline
$\beta$ & $\epsilon_1$ & $\epsilon_2$ & $l_{hit}$ & $l_{miss}$ & $\lambda_1$ & $\lambda_2$ & $\lambda_3$ & $\lambda_4$ & $\lambda_5$ \\ \hline
 1.0 & 1.0 & 1.2 & 0.4 & 0.05 & 0.2 & 0.9 & 1.0 & 0.1 & 0.4 \\ \hline
\end{tabular}
\label{Tabel:Coefficients}
\end{table}

\begin{figure*}
\centerline{\psfig{figure=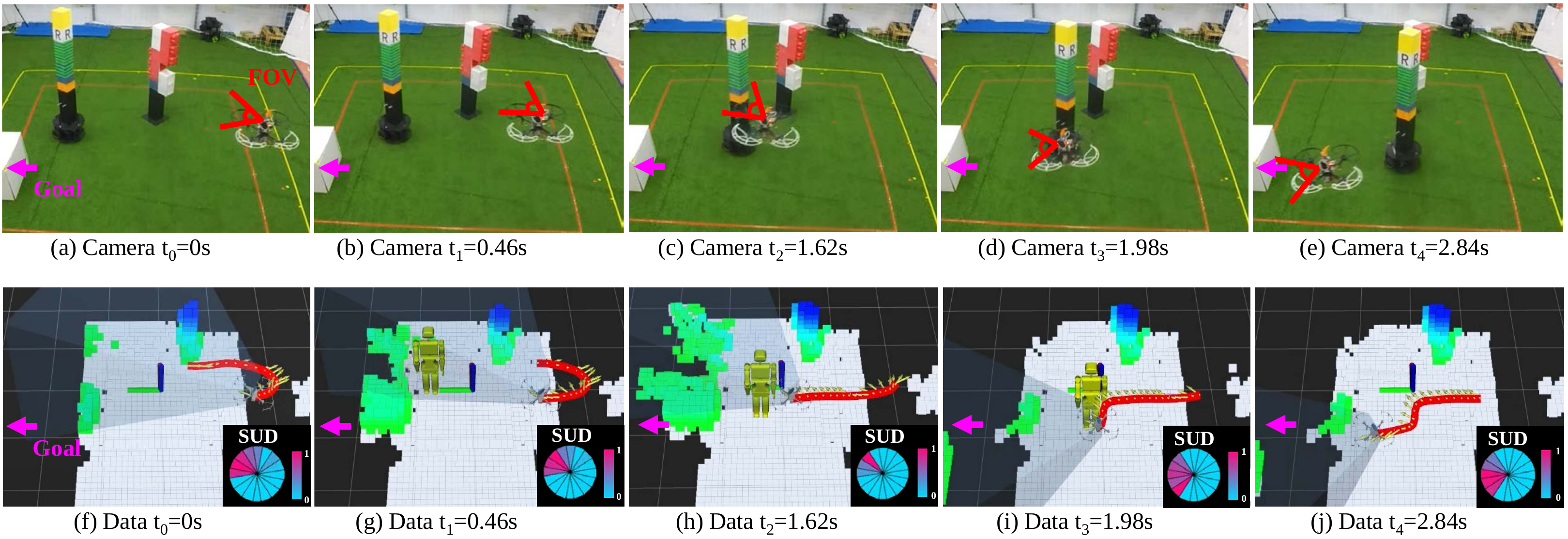,width=6.7in}}
\caption{Episode 1: collision avoidance to a dynamic obstacle. (a) to (e) present bird's-eye view snapshots, where the violet arrow or sphere indicates the goal position and the red angle marker presents the horizontal FOV. (f) to (j) illustrate the data visualization in RViz. Static obstacles are described as colored point clouds, while the dynamic obstacle is described with a yellow robot model. The FOV is shown by a translucent pyramid, and the trajectory in the last three seconds is presented with a red curve. Arrows on the curve indicate the direction of the camera. \hl{The rounded color maps in the bottom right corner illustrate the SUD of different directions.}}
\label{OneRobot}
\end{figure*}

\begin{figure*}
\centerline{\psfig{figure=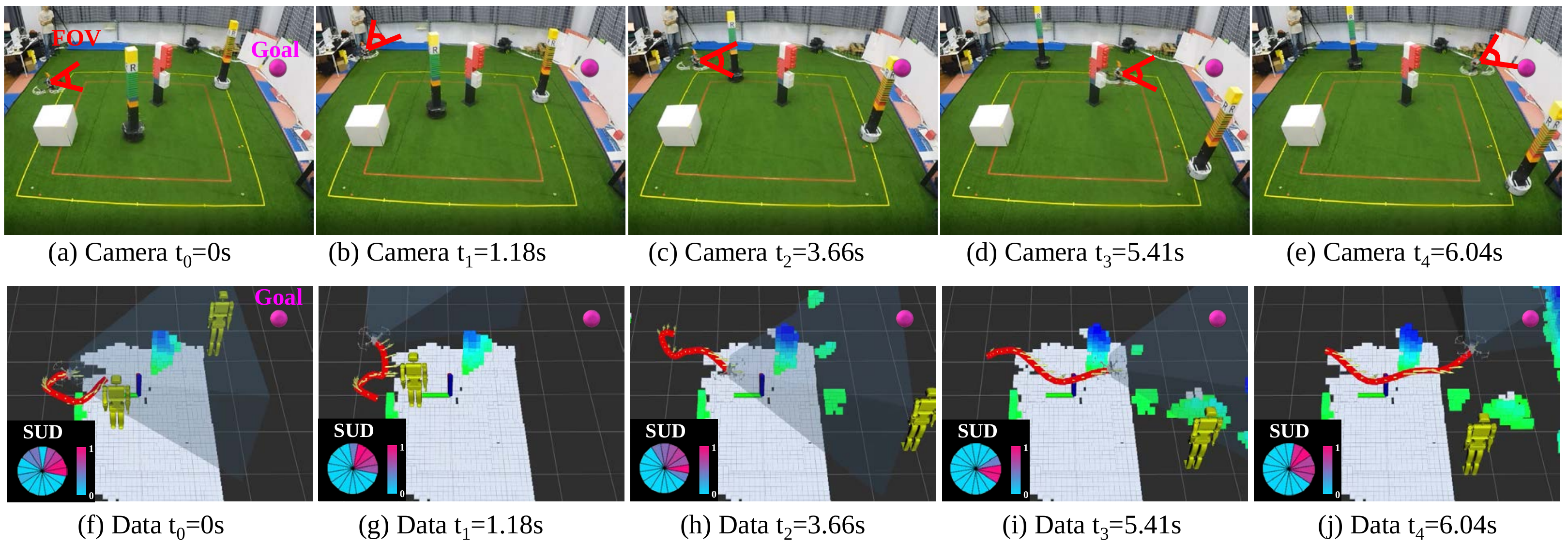,width=6.7in}}
\caption{Episode 2: abrupt goal direction change and collision avoidance to dynamic obstacles. (a) to (e) present bird's-eye view snapshots. (f) to (j) illustrate the data visualization in RViz. \hl{The rounded color maps in the bottom left corner illustrate the SUD of different directions.} The rest symbols are the same as the symbols in Fig. \ref{OneRobot}.}
\label{TwoRobots}
\end{figure*}

\begin{table}[]
\caption{Algorithms Benchmarking}
\centering
\begin{tabular}{l|lll}
\hline
Component & Detection \cite{Yolov3} & MOT & Map Building \cite{Ringbuffer} \\ \hline
Time & 60.5$\pm$8.6 ms$^*$ & 0.7$\pm$0.4 ms & 49.4$\pm$35.5 ms \\ \hline
Component & SUD & Camera Planning &  Path Planning \\ \hline
Time & 3.1$\pm$0.3 $\mu$s & 19.7$\pm$2.9 $\mu$s & 52.3$\pm$8.8 ms  \\ \hline  
\end{tabular}
\label{Tabel:TimeConsumption}
\end{table}

\begin{figure}
\centerline{\psfig{figure=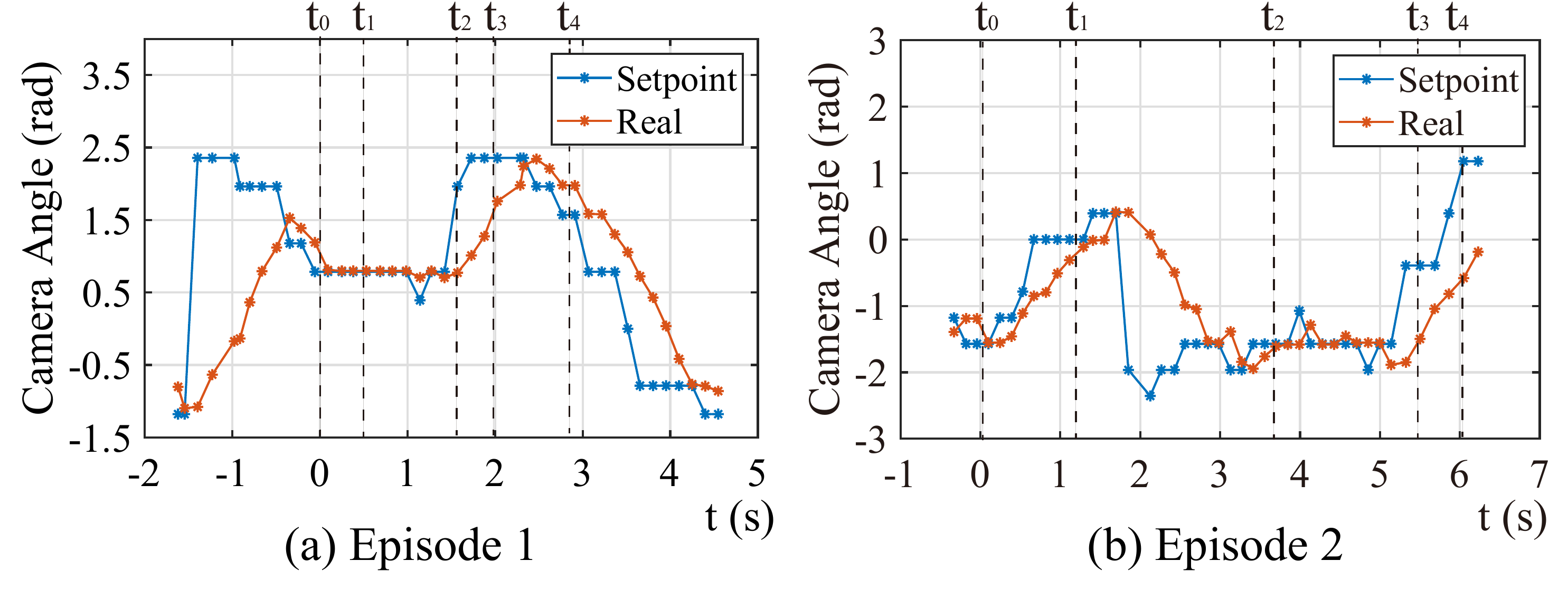,width=3.4in}}
\caption{\hl{Camera rotation angle curves of the two Episodes.}}
\label{camera_angle}
\end{figure}

\begin{figure}
\centerline{\psfig{figure=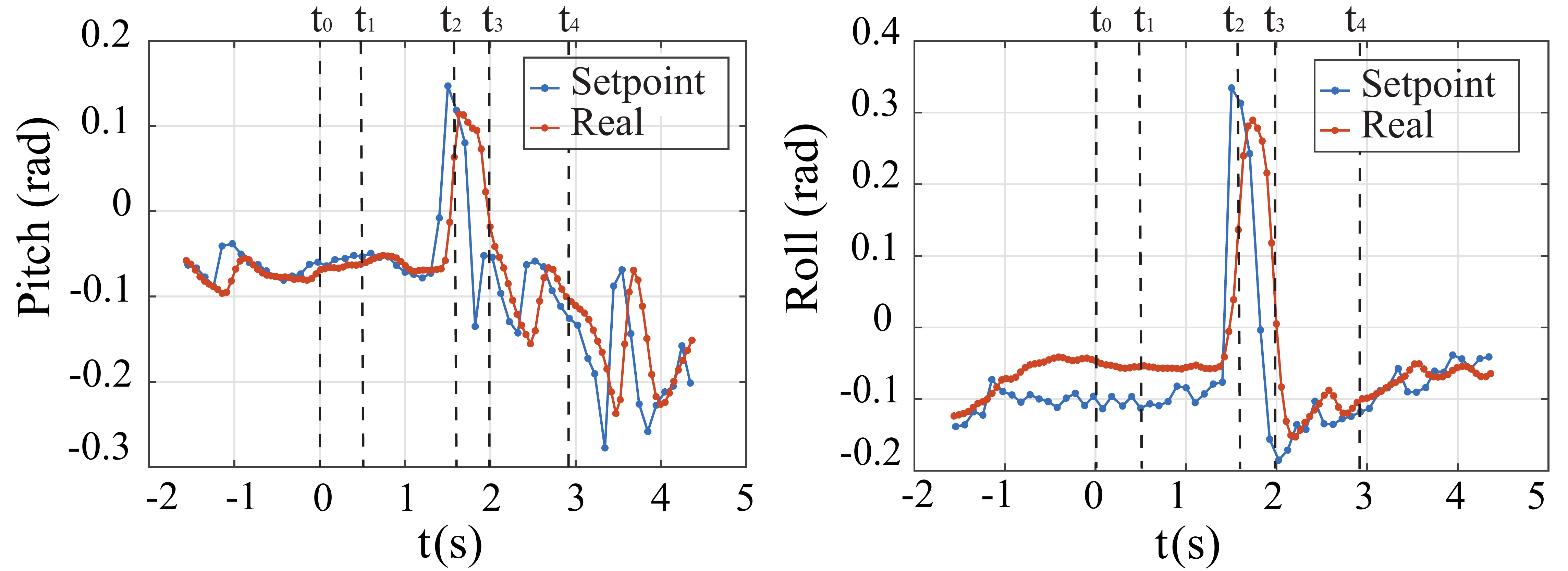,width=3.4in}}
\caption{Pitch and roll angle tracking curves of Episode 1.}
\label{One robot angle}
\end{figure}

\begin{figure}
\centerline{\psfig{figure=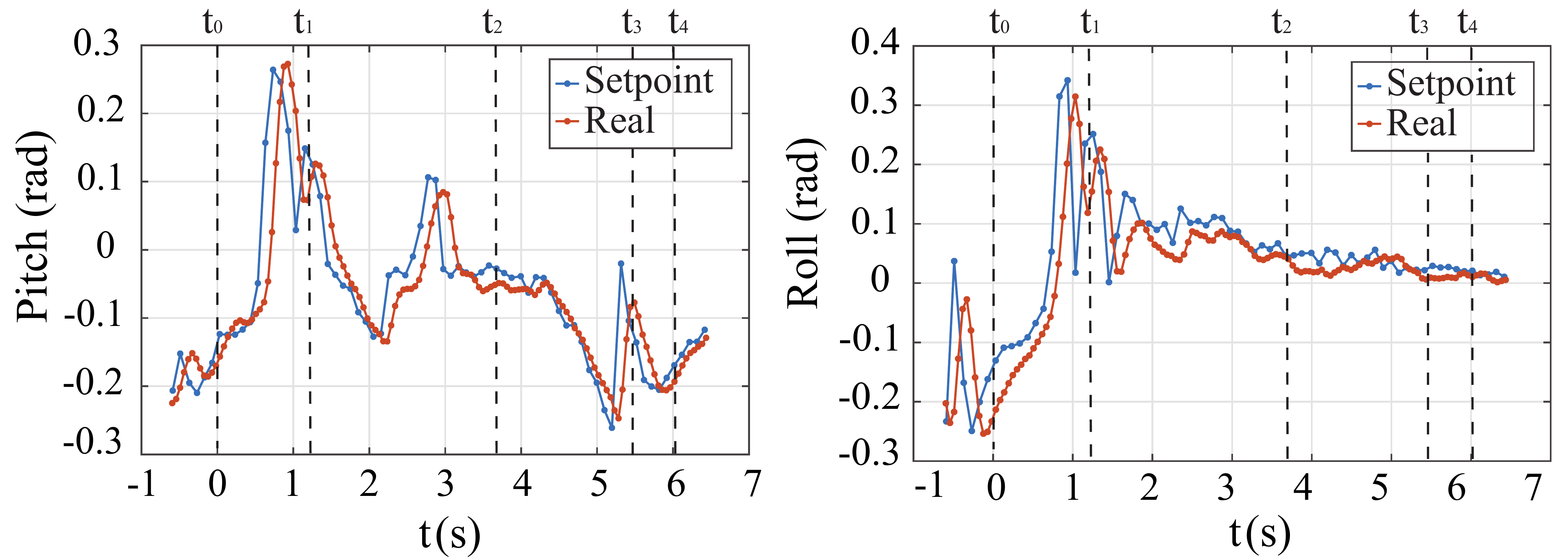,width=3.4in}}
\caption{Pitch and roll angle tracking curves of Episode 2.}
\label{two robots angle}
\end{figure}

We first measured the time consuming of the utilized algorithms in our onboard computer. The results are presented in Table \ref{Tabel:TimeConsumption}.
The star on the time consumption of dynamic obstacle detection indicates that the delay caused by this process is different. Unlike other algorithms in our system, detection is running in parallel on three neural computing chips. The delay from an image is captured to the detection is finished is as large as 180 ms. Thus the prediction by the model in the KF is necessary. The time consumption of updating SUD and planning camera motion is within 30 $\mu$s. Thus the increase in computing requirements when adding an independent DOF to realize active stereo vision is very little.

Fig. \ref{MAV}(b) presents a picture synthesized from three frames, in which the flying robot passes through the gap between a dynamic obstacle and a static obstacle rapidly. Two more typical episodes during the test are analyzed as follows.

a) Episode 1:
starts with a situation when the flying robot received a new goal position and turned to fly to the left side of the field. Five snapshots with time stamps and the corresponding map views are presented in Fig. \ref{OneRobot}. \hl{The camera rotation angle plot can be found in Fig. \ref{camera_angle}(a), and the corresponding pitch angle plot and roll angle plot are shown in Fig. \ref{One robot angle}. The yaw angle is not presented because it remained zero during the whole test. The timestamp of each snapshot is marked in the plots. The rounded color maps in the bottom right corner of Fig. \ref{OneRobot}(f) to Fig. \ref{OneRobot}(j) illustrate the SUD of different directions. It can be seen that from $t_1$ to $t_2$, the stereo camera was tracking the dynamic obstacle to predict its trajectory and the rotation angle had only a slight change. The SUD of the directions out of the FOV decreased. At $t_2$, the SUD of the left direction was smaller than the SUD of the top-left direction though these two directions were all included in the FOV. The reason is that the flying robot moved to the left during this period and brought the unknown area in this direction.}
From $t_2$ to $t_3$, an agile maneuver was taken to avoid collisions. For the reason that $l_{vis}$ is intentionally set to be quite small in our path planner, this maneuver was only taken when the flying robot was close to the obstacle. Pitch angle and roll angle in Fig. \ref{One robot angle} both reached a peak at $t_2$ to decelerate and make a turn. The red curve in Fig. \ref{OneRobot}(j) presents the flight trajectory of this episode, and the yellow arrows on the red curve indicate the rotation angle of the stereo camera.

b) Episode 2:
starts with a situation when the flying robot received a new goal position and turned to fly to the right side. However, a dynamic obstacle came in a sudden and blocked its way (Fig. \ref{TwoRobots}(a) and \ref{TwoRobots}(f)). Thus an agile maneuver was taken, and a motion primitive to the forward (in camera view, which is posterolateral to the goal direction) was chosen temporarily to find another valid path to the goal. \hl{Referring to Fig. \ref{camera_angle}(b), the camera rotated to the direction of this motion primitive and then turned back to observe the dynamic obstacle.} Unfortunately, the dynamic obstacle also moved forward and continued to block the way (Fig. \ref{TwoRobots}(b) and \ref{TwoRobots}(g)). According to the pitch angle plot in Fig. \ref{two robots angle}, the flying robot turned back at about $t=1.5 s$ because of the obstruction of the wall.
Then at $t_2$, a collision-free motion primitive to the right was successfully found since the dynamic obstacle moved away.
Finally, from $t_2$ to $t_4$, the flying robot successfully avoided collision with the static obstacle and approached the goal.
The dynamic obstacle on the right side was also detected and tracked while it moved far from the flight path and had little effect on the flight path.

During the test, the camera in our ASAA system rotated flexibly according to our optimizer's planning result to observe the neighborhood area and prevent the collision. \hl{In Fig. \ref{OneRobot}(i) and \ref{TwoRobots}(g), the dynamic obstacle was out of the FOV but was still considered in collision checking with the prediction result. When the dynamic obstacle had not been detected for two seconds, the covariance was too large, and thus the prediction was no longer utilized (Fig. \ref{OneRobot}(j) and Fig. \ref{TwoRobots}(h)).} More experiment results are presented in the video: \url{https://youtu.be/VohL2d_yYpg}.


\section{Conclusion}
Obstacle avoidance for flying robots in unknown and dynamic environments is a task of great challenge.
This paper presents a novel sense and avoid system using active stereo vision to tackle this task.
Comparison experiments in simulations have demonstrated the advantage of this system over the traditional fixed-camera systems. Real-world experiments also validated the effectiveness of this system in regard to dynamic obstacles and abrupt goal direction changes.
Unlike many existing flying robots that try to fuse multiple cameras to realize obstacle avoidance in dynamic environments, our system requires only one stereo camera, which can substantially reduce cost and benefit commercialization.
In future works, we will study more on biological behaviors and integrate reinforcement learning-based methods to further improve the active sensing performance. 


\bibliographystyle{IEEEtran}

\bibliography{head}

\section*{Acknowledgement}
This work is partially supported by the Scientific and Technical Innovation 2030 - “Artificial Intelligence of New Generation” Major Project (2018AAA0102704) and National Natural Science Foundation of China (Grant No. 51975348, 51605282). The object detection model training is also partially supported by grants from NVIDIA cooperation.

%
\begin{IEEEbiography}[{\includegraphics[width=1in,height=1.25in,clip,keepaspectratio]{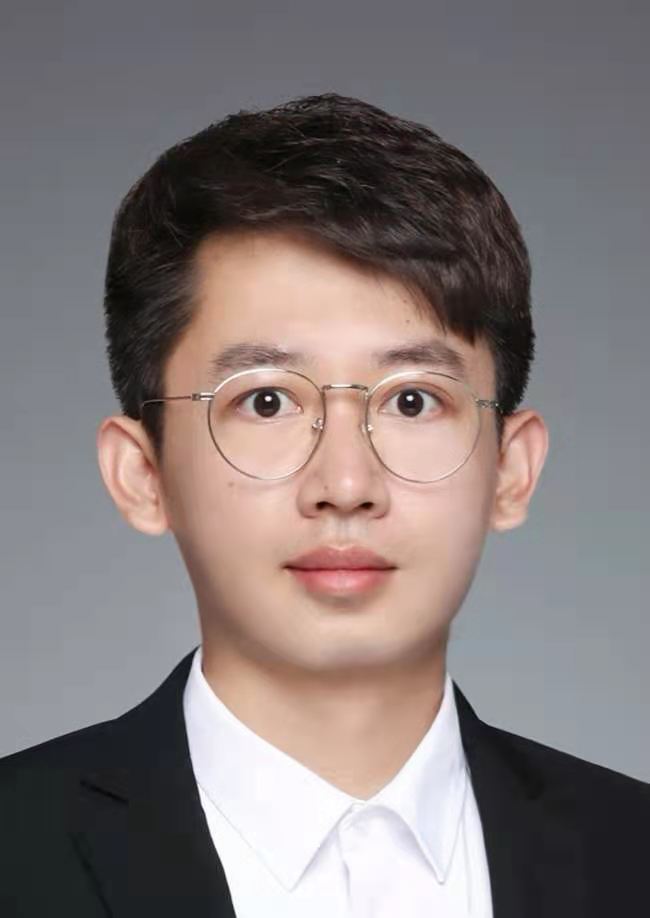}}]{Gang Chen}
Gang Chen received the B.E. degree in mechanical engineering from Shanghai Jiao Tong University, Shanghai, China, in 2016. He is now a Ph. D. candidate at the State Key Laboratory of Mechanical System and Vibration, School of Mechanical Engineering, Shanghai Jiao Tong University. His research interest is on vision-based obstacle avoidance for robotics system.
\end{IEEEbiography}
\begin{IEEEbiography}[{\includegraphics[width=1in,height=1.25in,clip,keepaspectratio]{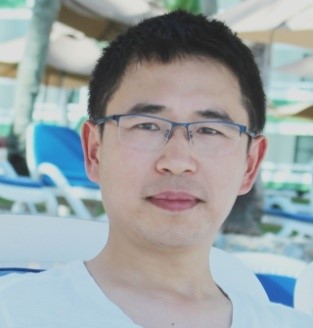}}]{Wei Dong}
Wei Dong received the B.S. degree and Ph.D. degree in mechanical engineering from Shanghai Jiao Tong University, Shanghai, China, in 2009 and 2015, respectively. He is currently an assistant professor in the School of Mechanical Engineering, Shanghai Jiao Tong University. His research interests include cooperation, perception and agile control of unmanned system.
\end{IEEEbiography}
\begin{IEEEbiography}[{\includegraphics[width=1in,height=1.25in,clip,keepaspectratio]{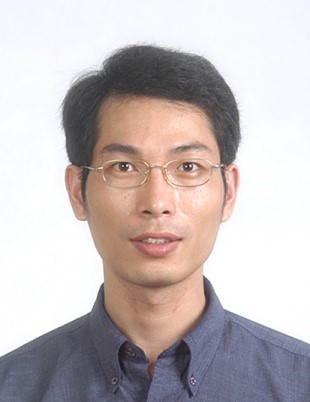}}]{Xinjun Sheng}
Xinjun Sheng received the B.Sc., M.Sc.,and Ph.D. degrees in mechanical engineering fromShanghai Jiao Tong University, Shanghai, China, in2000, 2003, and 2014, respectively. In 2012, he was a Visiting Scientist at Concordia University, Montreal, QC, Canada. He is currently an associate professor in the School of Mechanical Engineering,Shanghai Jiao Tong University. His current research interests include robotics, and bio-mechatronics. Dr. Sheng is a Member of the IEEERAS, the IEEEEMBS,and the IEEEIES.
\end{IEEEbiography}
\begin{IEEEbiography}[{\includegraphics[width=1in,height=1.25in,clip,keepaspectratio]{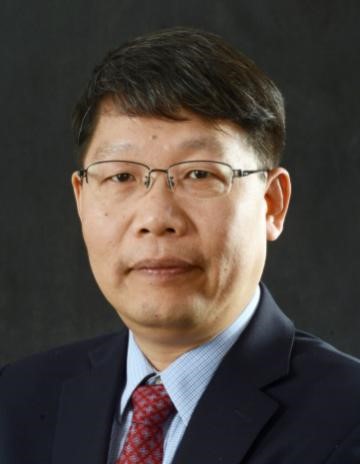}}]{Xiangyang Zhu}
Xiangyang Zhu received the B.S. degree from the Department of Automatic Control Engineering, Nanjing Institute of Technology, Nanjing, China, in 1985, the M.Phil. degree in instrumentation engineering and the Ph.D. degree in control engineering, both from Southeast Engineering, in 1989 and 1992, respectively. From 1993 to 1994, he was a postdoctoral research fellow with Huazhong University of Science and Technology, Wuhan, China. He joined the Department of Mechanical Engineering, Southeast University, as an associate professor in 1995. Since June 2002, he has been with the School of Mechanical Engineering, Shanghai Jiao Tong University, Shanghai, China, where he is currently a chair professor and the director of the Robotics Institute. His research interests include robotic manipulation planning, neuro-interfacing and neuro-prosthetics, and soft robotics. He has published more than 200 papers in international journals and conference proceedings.
Dr. Zhu has received a number of awards including the National Science Fund for Distinguished Young Scholars from NSFC in 2005, and the Cheung Kong Distinguished Professorship from the Ministry of Education in 2007. He is serving on the editorial board of IEEE Transactions on Cybernetics and Journal of Bionic Engineering.
\end{IEEEbiography}
\begin{IEEEbiography}[{\includegraphics[width=1in,height=1.25in,clip,keepaspectratio]{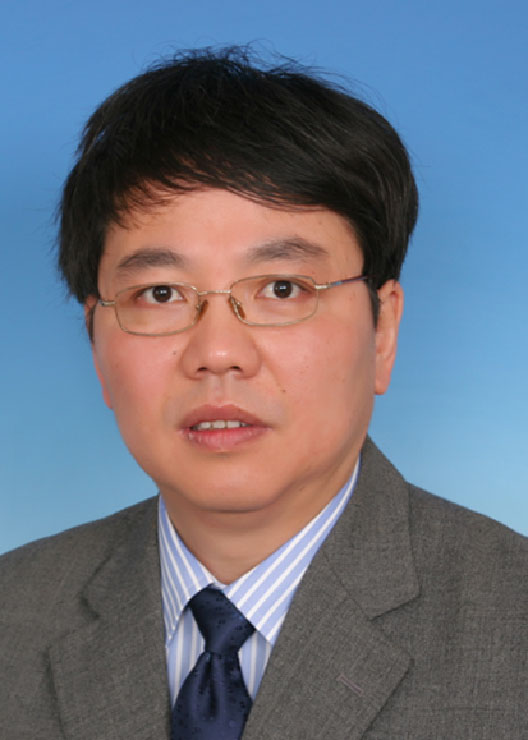}}]{Han Ding}
Han Ding received his Ph.D. degree from Huazhong University of Science and Technology (HUST), Wuhan, China, in 1989. Supported by the Alexander von Humboldt Foundation, he was with the University of Stuttgart, Germany from 1993 to 1994. He worked at the School of Electrical and Electronic Engineering, Nanyang Technological University, Singapore from 1994 to 1996. He has been a Professor at HUST ever since 1997 and is now Director of State Key Lab of Digital Manufacturing Equipment and Technology there.  Dr. Ding was a “Cheung Kong” Chair Professor of Shanghai Jiao Tong University from 2001 to now.  Dr. Ding acted as an Associate Editor of IEEE Trans. on Automation Science and Engineering(TASE) from 2004 to 2007. Currently, he is an Editor of IEEE TASE and a Technical Editor of IEEE/ASME Trans. on Mechatronics. His research interests include robotics, multi-axis machining and control engineering.
\end{IEEEbiography}

%
%
%
%
%
%
%
%
%

\end{document}